\documentclass{llncs}
\usepackage{times}
\usepackage{epsfig}
\usepackage{graphicx}
\usepackage{algorithm}
\usepackage{algpseudocode}
\usepackage{amsmath}
\usepackage{amssymb}
\usepackage{mathtools}
\DeclarePairedDelimiter{\ceil}{\lceil}{\rceil}
\usepackage{subfigure}
\usepackage{caption} 
\usepackage{enumitem}
\usepackage[dvipsnames]{xcolor}
\usepackage{setspace}
\begin{document}
\frontmatter          
\pagestyle{headings}  
\mainmatter              
\title{PatchFCN for Intracranial Hemorrhage Detection}

\author{Weicheng Kuo\inst{1} \and
Christian H{\"a}ne\inst{1}\and
Esther Yuh \inst{2} \and
Pratik Mukherjee \inst{2} \and
Jitendra~Malik\inst{1}}


%
\institute{University of California Berkeley, Berkeley CA, 94720, USA,\\
\and
University of California San Francisco School of Medicine,\\
San Francisco CA, 94143, USA,\\
\email{wckuo@berkeley.edu}}

\maketitle              
\vspace{-8mm}
\begin{abstract}
This paper studies the problem of detecting and segmenting acute intracranial hemorrhage on head computed tomography (CT) scans. We propose to solve both tasks as a semantic segmentation problem using a patch-based fully convolutional network (PatchFCN). This formulation allows us to accurately localize hemorrhages while bypassing the complexity of object detection. Our system demonstrates competitive performance with a human expert and the state-of-the-art on classification tasks ($0.976$, $0.966$ AUC of ROC on retrospective and prospective test sets) and on segmentation tasks ($0.785$ pixel AP, $0.766$ Dice score), while using much less data and a simpler system. In addition, we conduct a series of controlled experiments to understand \textit{why} PatchFCN outperforms standard FCN. Our studies show that PatchFCN finds a good trade-off between batch diversity and the amount of context during training. These findings may also apply to other medical segmentation tasks.
 
\end{abstract}
\vspace{-8mm}
\section{Introduction}
\vspace{-2mm}
Traumatic brain injury (TBI) is a major contributor to injury-related deaths. In emergency departments, head computed tomography (CT) scans are routinely performed on patients under evaluation for suspected TBI. Existing works have shown that a computer vision system that rapidly and reliably detects emergency TBI findings, such as acute intracranial bleeding, on head CT scans can significantly reduce the time to diagnosis and potentially reduce death and long-term disability \cite{arbabshirani2018advanced,titano2018automated}. Deep learning techniques have been successful recently in detecting intracranial hemorrhages, e.g. 3D classification \cite{arbabshirani2018advanced,titano2018automated} supervised by text reports, 2D classification \cite{lee2019explainable}, instance segmentation \cite{chang2018hybrid}. However, to our knowledge, no semantic segmentation approach has shown performance competitive with human experts.

\begin{figure}[h]
    \centering
    \includegraphics[width=1.0\linewidth]{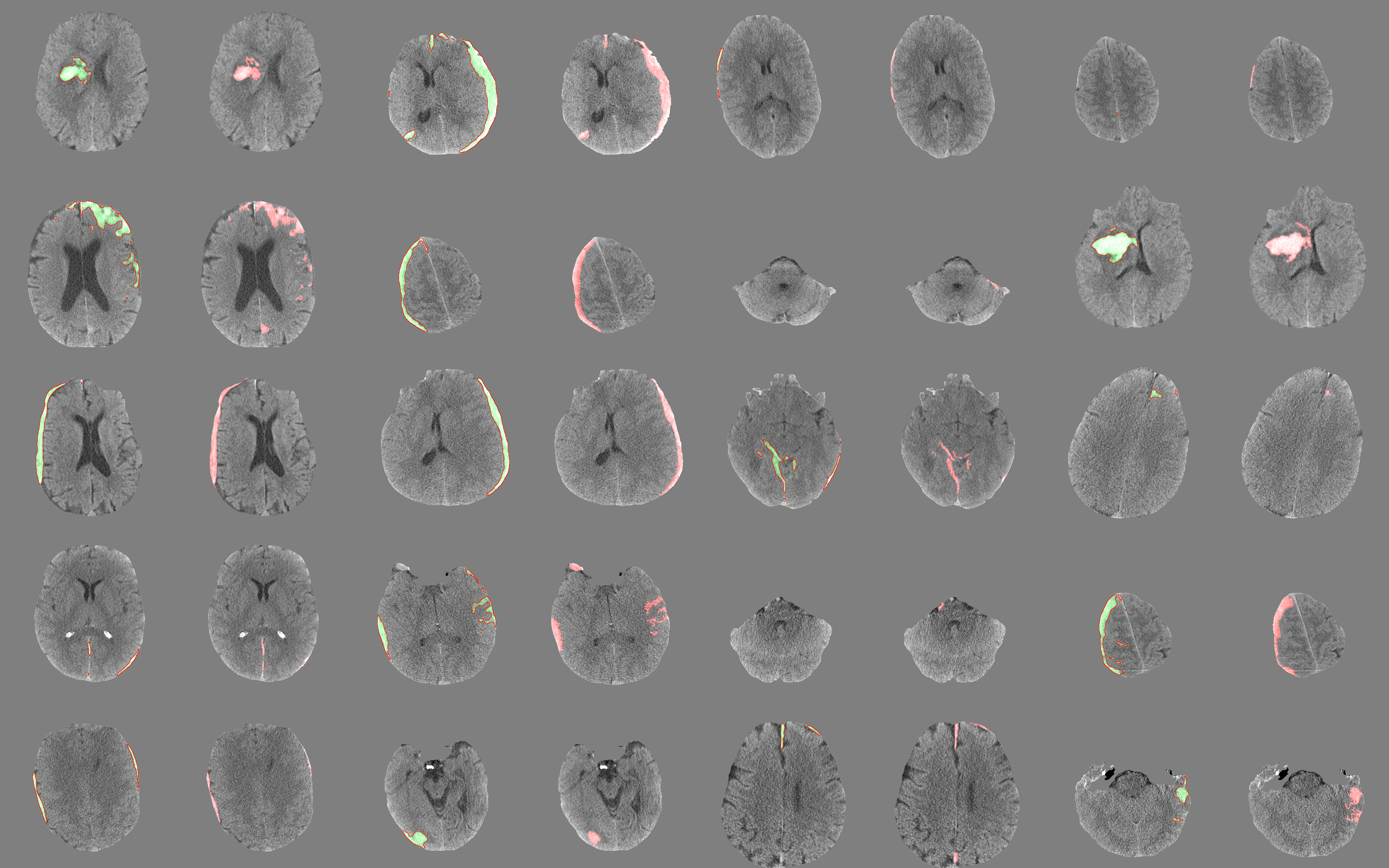}
    \caption{Visualization of PatchFCN segmentation. Each pair contains the PatchFCN output (left) and groundtruth labels (right). Results are randomly selected from the positive frames of the test set.}
    \vspace{-6mm}
    \label{fig:vis-seg}
\end{figure}

We propose to solve the detection and segmentation problem \textit{jointly} as a semantic segmentation task. Segmentation offers many advantages over classification, including better interpretability, and quantifiable metrics for disease prognosis \cite{chang2018hybrid,lee2019explainable}. Unlike \cite{chang2018hybrid}, we view hemorrhage as ``stuff'' (e.g. water) rather than ``things'' (e.g. car) due to its fluid nature. As the clinical need is to know whether a scan (i.e.\ whole head) is positive, and where the positive pixels are, semantic segmentation is the simplest way to achieve it. 

Among existing pixel-wise labeling techniques, fully convolutional networks \cite{long2015fully} (FCN) are successful and widely adopted for such tasks in computer vision \cite{long2015fully} and the medical imaging community \cite{zhang2018deep,qin2018autofocus}. Most computer vision practitioners use whole images as inputs for their FCNs following \cite{long2015fully}. This is in contrast to how patch-based FCN training has been successful in applications such as retinopathy \cite{zhang2018deep}, MRI \cite{qin2018autofocus}, and X-ray/CT imaging \cite{wang2017multi,zhang2018convolutional}. Despite the wide adoption, there exists no systematic study on why patches improve FCN in many cases.

We propose PatchFCN and show that it outperforms standard FCN in localizing hemorrhages. Since no public dataset is available, one important challenge we face is to acquire pixelwise labeled data. Unlike the approaches that learn from text reports \cite{arbabshirani2018advanced,titano2018automated}, we collect a dataset of $591$ scans annotated \textit{pixelwise} for the presence of hemorrhage by expert radiologists. Using $100$x smaller data, PatchFCN significantly outperforms weakly supervised methods \cite{arbabshirani2018advanced,titano2018automated} on classification tasks. Compared to the state-of-the-art segmentation method \cite{chang2018hybrid}, our segmentation and classification results are competitive while using $20$x less training data and a simpler system.

We analyze the following factors to better understand the performance gains of PatchFCN: 1) batch diversity, 2) amount of context, and 3) sliding window inference. We find that PatchFCN outperforms FCN by finding an optimal trade-off between batch diversity and the amount of context. In addition, sliding window inference helps to bridge the gap of train/test time and consistently improve performance. We hope these findings would benefit other segmentation tasks where patch-based training is effective.

\vspace{-2mm}
\section{Method}
The goals for hemorrhage detection are to find out: 1) whether a stack contains hemorrhage, and 2) where the hemorrhage is within the stack. In practice this may be used by the radiologists/neurosurgeons to assess the risk level of the patient and triage the patient to immediate surgical evacuation, monitoring in the intensive care unit (ICU), or routine monitoring on the hospital ward. Inspired by existing works \cite{qin2018autofocus,zhang2018deep,wang2017multi,zhang2018convolutional,karayumak2018harmonizing}, we propose to solve both tasks with PatchFCN as follows (see Fig.\ref{fig:patch}):
\vspace{-3mm}
\begin{figure}[t]
    \centering
    \includegraphics[width=0.5 \linewidth]{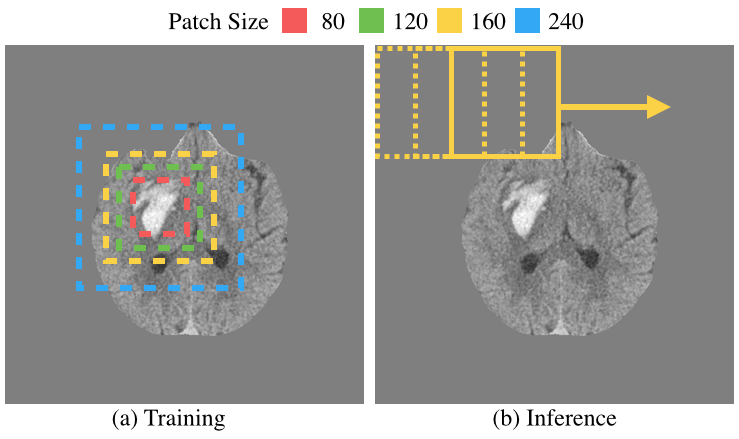}
    \caption{PatchFCN train on small patches and test in sliding window fashion. The colored boxes show different patch sizes in the context of a hemorrhage.}
    \vspace{-6mm}
    \label{fig:patch}
\end{figure}

\subsubsection{Patch-based Training:} We train an FCN on random small patches cropped from the whole images centered on foreground. The model learns to predict the binary pixel label within the patches. For head CT data, the intuition of patch-based training comes from how radiologists make decisions -- the morphology of contrast region is often a crucial cue for deciding whether it represents pathologies. Similarly, PatchFCN causes the network to make its decision based on the local image information without relying on excessive context. In addition, small patches allow larger batch size and hence higher batch diversity to stabilize network training. As most convolutional networks have built-in batch normalization e.g. \cite{yu2017dilated}, PatchFCN leverages it by finding a good trade-off between large minibatch and adequate context for the task.
\vspace{-2.5mm}
\subsubsection{Patch-based Inference:} At test time, we evaluate the images in a sliding window fashion, as opposed to the typical fully convolutional inference. Sliding window at test time avoids any domain shift which occurs when training on small patches and evaluating fully convolutionally on the whole image. This is because the paddings present in convolution layers make a patch in the context of a whole image not the same as the patch by itself. Let the input image be of size $H$ and the patch size $C$, then the total number of windows is given by $N=\ceil{\frac{\beta H}{C}}^2$, where $\beta > 1$ is an adjustable parameter for the window overlap. As multiple predictions are made for each pixel, we simply average their scores. The frame-level score is obtained by averaging the pixel scores within the frame. To get stack-level scores from pixel scores, we first take $L^p$-norm over the frame to obtain a stack-frame score. The stack score is defined as the maximum stack-frame score within a stack. $p$ is treated as a hyper-parameter and tuned on the trainval set.
\vspace{-2.5mm}
\subsubsection{Data Collection:} Our dataset consists of $591$ clinical head CT scans performed over 7 years from 2010 to 2017 on 4 different 64-detector-row CT scanners (GE Healthcare, Siemens) at our affiliated hospitals. We use the word ``stack" for each patient's head CT scan, and the word ``frame" for each individual slice of the stack. The scans were anonymized by removing patient-related meta-data, skull, scalp, and face. Board-certified senior neuroradiologists who specialize in TBI identified all areas of hemorrhage in our dataset. 
Our data contains the typical spectrum of technical limitations seen in clinical practice (e.g. motion artifact, ``streak'' artifact near the skull base or in the presence of metal), and also contains all of the subtypes of acute intracranial hemorrhage, including epidural hematoma, subdural hematoma, subarachnoid hemorrhage, hemorrhagic contusion, and intracerebral hemorrhage (see Fig.~\ref{fig:benchmark} for examples). We randomly split the data into a trainval/test set of $443$/$148$ stacks for development and internal validation. The hyper-parameters of PatchFCN are tuned within the trainval set.

\vspace{-4mm}
\subsubsection{Implementation Details:} We choose a DRN-38 backbone because it performs competitively among many network designs \cite{yu2017dilated}. Regarding the inputs, we clip the dynamic range of raw data at $-40$ and $90$ Hounsfield unit (HU), and then rescale the intensity to lie within $[0,255]$. Image size is $512 \times 512$. In both training and test time, we use a patch size of $240$ unless stated otherwise. We utilize the z-axis context by fusing the adjacent frames with the center frame at the input (3 channels in total). The optimization is done by SGD with momentum following \cite{yu2017dilated} setup. We train the network from scratch without using ImageNet pretraining, as we do not observe any gains using ImageNet. We re-weight the positive class loss by $\alpha=3$ to balance the dominant negative class loss. The learning rate starts at $0.005$ and decreases by a factor of $0.1$ after $40\%$ and $80\%$ of the complete training iterations. At test time, we select $\beta=3$ to ensure good overlap between adjacent sliding windows. To compute stack-level score, we select $p=256$ in the $L^p$ norm. All parameters were found by cross validation on the trainval set.

\vspace{-4mm}
\section{Experiments}
\vspace{-2mm}
\subsection{Stack-level Benchmark with Human Experts}
\begin{figure}[t]
    \centering
    \includegraphics[width=0.45\linewidth]{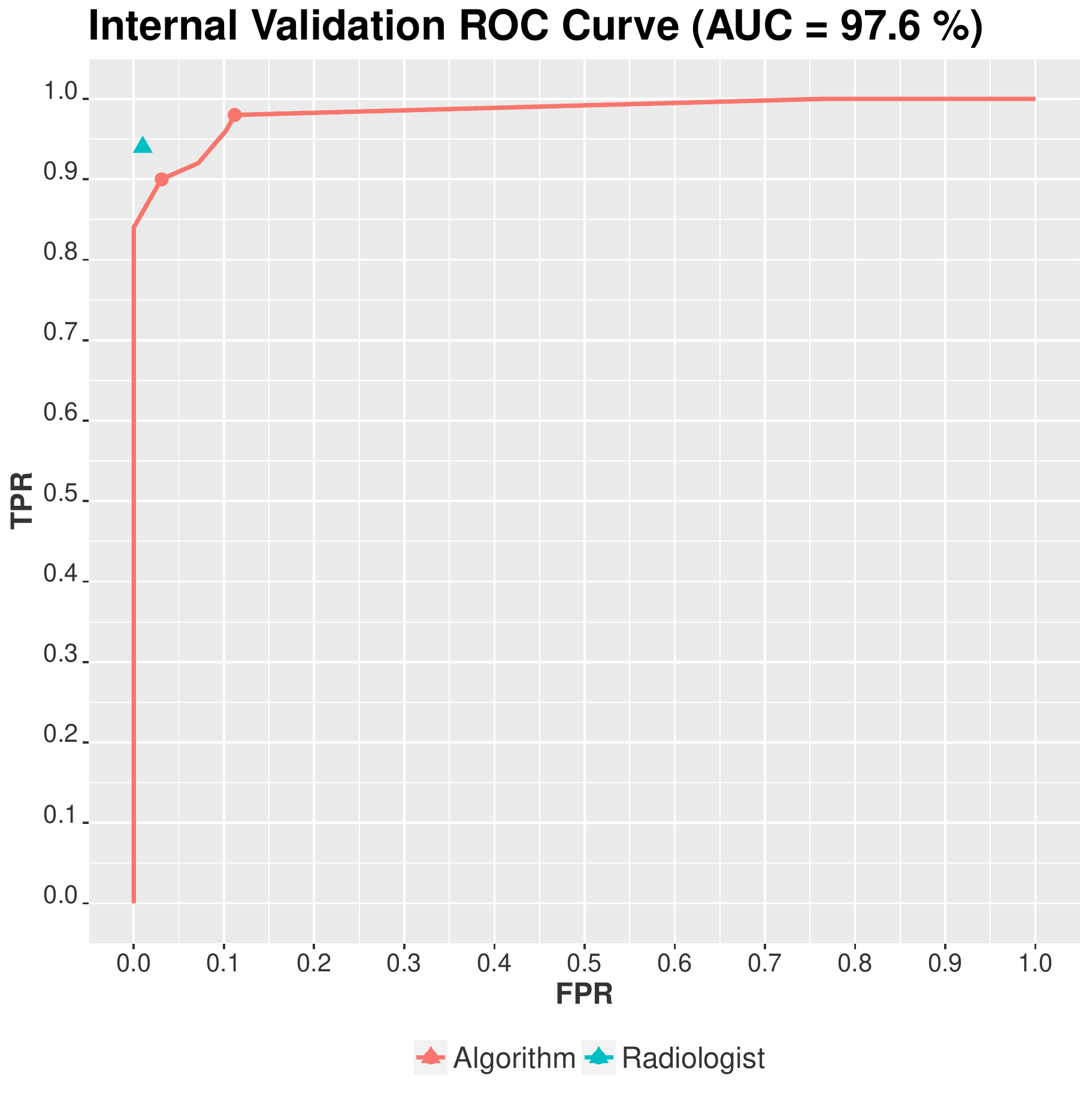}
    \includegraphics[width=0.45\linewidth]{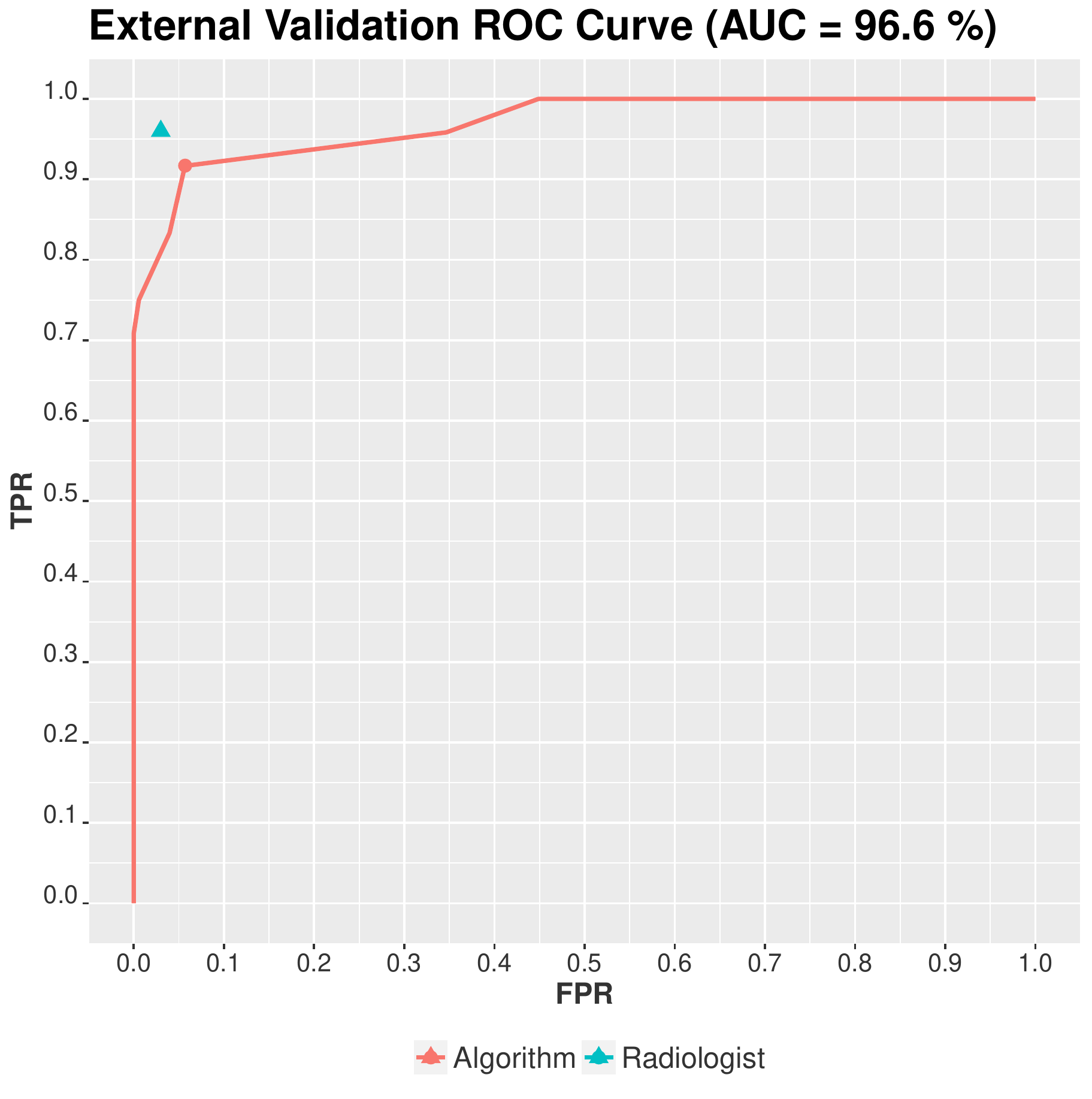}
    \vspace{-3mm}
    \caption{Internal and External Validation. We compare PatchFCN to an expert (neuroradiology attending with 15 years of experience) at stack level on retrospective and prospective test sets. PatchFCN achieves AUCs of $0.976$ and $0.966$ respectively, competitive with state-of-the-art systems that use much more labeled data. PatchFCN approaches but does not exceed the attending neuroradiologist.}
    \vspace{-2mm}
    \label{fig:benchmark}
\end{figure}

The first order task of hemorrhage detection is to determine whether a stack contains hemorrhage. We conduct internal as well as external validation for PatchFCN on stack-level as shown in Figure \ref{fig:benchmark}. The human expert is a neuroradiologist certified by the American Board of Radiology with 15 years of attending experience. The expert is instructed to examine each scan with the same level of care as a clinical scan. We allow the expert to take as much as time as needed. The expert can modify their reads on scans before submitting final answers on the whole data set. The groundtruths are determined by at least one neuroradiologist with more than 10 years of neuroradiology attending experience.
\vspace{-3mm}
\subsubsection{Internal (Retrospective) Validation:} We report the ROC curve of PatchFCN on the test set and compare it with a human expert (15-year attending) in a retrospective setting where the test data was collected before the model development. Our single model AUC of $0.976$ is competitive against the state-of-the-art $0.983$ (single model) \cite{chang2018hybrid} and $0.993$ (ensemble models) \cite{lee2019explainable}, while using much less training data. Our human expert has very low false positive rate $0.01$ at $0.94$ recall, better than the $(0.03, 0.90)$ of PatchFCN. Using both trainval and test data, our 4-fold cross validation AUC is $0.971 \pm 0.006$.

\vspace{-3mm}
\subsubsection{External (Prospective) Validation:} We collected a prospective test set of $200$ scans after the model was developed. No further hyper-parameter adjustment was allowed in order to prevent overfitting to the test set. To minimize selection bias, we randomly select from all head CT scans performed from November to December 2018 using the Radiology Information System (RIS) SQL database in our hospital. The positive rate is $12.5\%$, which approximates the observed positive rates in emergency departments of many U.S. hospitals. Our ensemble model ($n=3$) achieves an AUC of $0.966$, which is competitive against the state-of-the-art $0.981$ \cite{chang2018hybrid} and $0.961$ \cite{lee2019explainable}. PatchFCN approaches but does not exceed the human expert. Our best operating point is $(0.06, 0.92)$.

\subsection{Pixel-level Evaluation}
\vspace{-2mm}
Apart from stack-level evaluation, we evaluate PatchFCN at pixel level because clinicians also want to know the location and volume of the bleeds for disease prognosis. Figure \ref{fig:vis-seg} visualizes the outputs of PatchFCN in comparison with the groundtruths. Results are shown on randomly selected positive frames in the retrospective test set.

On the retrospective test set, our model achieves pixelwise Dice score, Jaccard index, and average precision of $0.766$, $0.620$, and $0.785$. In comparison, \cite{chang2018hybrid} reports Dice scores of $0.77$ to $0.93$ for a few types of hemorrhages they study. Our groundtruths are annotated pixelwise by senior neuroradiologists who specialize in TBI and include many subtle findings that could be easily missed by inexperienced radiologists. Using both trainval and test data, our 4-fold cross validation Dice score is $0.722 \pm 0.027$. 

\vspace{-3mm}
\begin{table}[t]
     \centering
     \def\arraystretch{1.1}\tabcolsep=6pt
     \begin{tabular}{|c | c c c c c |} 
     \hline
     Crop Size & 80 & 120 & 160 & 240 & 480 \\
     \hline
     Batch Size & 144 & 64 & 36 & 16 & 4 \\
     \hline
     Epoch & 3600 & 1600 & 900 & 400 & 100 \\
     \hline\hline
     Dice & 75.5 & 75.9 & 76.2 & 76.6 & 74.2 \\
     \hline
     Jaccard & 60.7 & 61.2 & 61.6 & 62.0 & 59.0 \\
     \hline
     Pixel AP & 78.5 & 78.1 & 78.5 & 78.5 & 75.9 \\
     \hline
     Frame AP & 87.8 & 89.3 & 89.8 & 89.9 & 87.8 \\
     \hline
    \end{tabular}
    \vspace{-1mm}
    \caption{We benchmark PatchFCN on different patch sizes. Patch size 480 is the standard FCN that consumes whole images (baseline). As seen, PatchFCN consistently outperforms the baseline across a wide range of patch sizes on pixel and frame metric.}
    \vspace{-8mm}
    \label{table:hemorrhage}
\end{table}

\subsection{PatchFCN vs. FCN}
\label{sec:patch_learn}

Table ~\ref{table:hemorrhage} shows that PatchFCN consistently improves over standard FCN for pixel and frame by a healthy margin for a wide range of patch sizes. We report average precision (AP), Dice score and Jaccard index at pixel level with a threshold of 0.5. 
Note how PatchFCN is robust to patch size and maintains the performance even at a patch size of 80. We have tried even smaller sizes and observed a significant performance drop due to difficult optimization. To compare across different patch sizes, we choose the batch size to control the number of input pixels per batch to be the same, and we choose the number of epochs such that the number of gradient steps are the same. We also ensure that all performances are saturated and training longer does not improve further.  

\vspace{-3mm}
\subsection{What Makes PatchFCN effective?}
\label{sec:patch_effective}
\vspace{-2mm}
Given the effectiveness of PatchFCN, we want to delve deeper to understand what makes patches so effective. We identify a few differences from standard FCN and study them by control experiments. For the following experiments, we define the batch size $B$, which is the product of $N$, the number of images per batch, and $K$, the number of patches per image.  The batch size is defined this way because we sample patches from each of the image samples. PatchFCN has $K=1$, $N=16$, $B=16$ and $C=240$, where $C$ is the crop size, whereas the standard FCN has $K=1, N=4, B=4, C=480$. We perform these analyses on the test split because it is larger and yields more stable performance. In this section, we control the number of input pixels and number of iterations the same way as in Section \ref{sec:patch_learn}, unless otherwise stated.

\vspace{-2mm}
\subsubsection{Batch Diversity:} 
\vspace{-2mm}
One possible advantage of PatchFCN is that we can fit a larger batch size and thus include more diverse data within any given GPU memory. To study the contribution of batch diversity, we control the batch size $B$ and decrease the number of images $N$ we sample patches from. Since $B=N \times K$, this means we sample more patches per image. As $N$ decreases, we expect batch diversity to decrease as well. The default PatchFCN has $N=B$ and $K=1$, which has the greatest diversity for any given $B$. By fixing the other hyperparameters, we can safely say the only difference here is the batch diversity. Note that we control the number of steps to be the same, so we decrease the number of epochs linearly with $N$.

Table \ref{table:batch_diversity} shows that decreased batch diversity results in lower pixel and frame-level performance. The breaking point is at $N=2$, where the performance drops significantly from $N=4$. We speculate that this is due to the use of batch normalization in residual networks\cite{yu2017dilated}. This experiment demonstrates the importance of batch diversity for PatchFCN.
\vspace{-4mm}
\begin{table}[t]
     \centering
     \def\arraystretch{1.1}\tabcolsep=6pt
     \begin{tabular}{| c c c c c | c c c c |} 
     \hline
     $N$ & $K$ & $B$ & $C$ & Epoch & Dice & Jaccard & PixelAP & FrameAP \\
     \hline\hline
     16 & 1 & 16 & 240 & 400 & 76.6 & 62.0 & 78.5 & 89.8 \\
     \hline
     8 & 2 & 16 & 240 & 200 & 76.4 & 61.8 & 78.5 & 89.7\\
     \hline
     4 & 4 & 16 & 240 & 100 & 74.7 & 59.6 & 77.3 & 87.7\\
     \hline
     2 & 8 & 16 & 240 & 50 & 57.5 & 40.3 & 67.6 & 81.4\\
     \hline
    \end{tabular}
    \caption{PatchFCN performance decreases with decreasing batch diveristy.} 
    \label{table:batch_diversity}
\end{table}

\subsubsection{How Much Context Does PatchFCN Need?}
A trade-off of using patches is that we restrict the amount of context available to the network during training. Intuitively, one would think that more context is better. However, with limited amount of data, it is possible that less context could serve as an effective regularizer by forcing the prediction to rely on local information. To understand how much we lose/gain by having less context, we compare PatchFCN using different patch sizes while fixing the batch size and the number of steps (number of input pixels not the same here). 

\begin{table}[t]
\vspace{-7mm}
     \centering
     \def\arraystretch{1.1}\tabcolsep=6pt
     \begin{tabular}{|c c c c c | c c c c |} 
     \hline
     $C$ & $N$ & $K$ & $B$ & Epoch & Dice & Jaccard & PixelAP & FrameAP \\
     \hline\hline
     64 & 16 & 1 & 16 & 400 & 66.4 & 49.7 & 65.8 & 74.5 \\
     \hline
     120 & 16 & 1 & 16 & 400 & 72.5 & 56.9 & 74.7 & 82.2 \\
     \hline
     240 & 16 & 1 & 16 & 400 & 76.6 & 62.0 & 78.5 & 89.9 \\
     \hline
     360 & 16 & 1 & 16 & 400 & 73.9 & 58.6 & 73.4 & 85.8 \\
     \hline
     480 & 16 & 1 & 16 & 400 & 74.1 & 58.8 & 75.6 & 87.7 \\
     \hline
    \end{tabular}
    \vspace{-1mm}
    \caption{Context helps PatchFCN from $C=64$ to $240$, but not beyond.}  \vspace{-5mm}
    \label{table:context}
\end{table}

Table \ref{table:context} shows that the improvement of context plateaus at patch size $C=240$. Compared to $C=64$, $C=240$ is significantly better. However, increasing the patch size beyond 240 does not offer any more gain. We speculate that the improvement comes from the context regularization of patches, which helps in case of limited data. Overall, controlling context with patches is effective and allows the use of a larger and more diverse batch as in Table \ref{table:batch_diversity}.

\begin{figure}[t]
    \centering
    \includegraphics[width=1.0\linewidth]{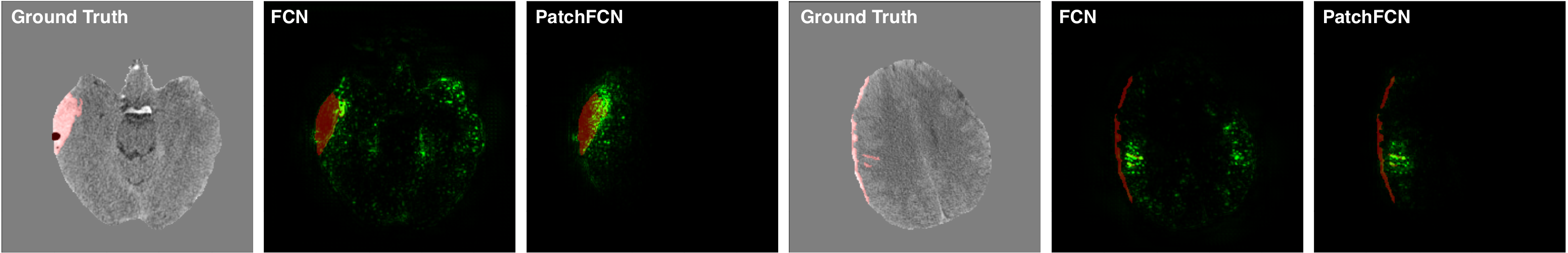}
    \caption{We visualize the gradients of PatchFCN with FCN in image space to see what cues the models rely on. Green speckles are the gradients and the burgundy regions are the selected ground truths for back-propagation.}
    \label{fig:grad_vis}
\end{figure}

To qualitatively study what cues PatchFCN uses, we backpropagate the gradients from each hemorrhage region to the image space (see Fig.\ref{fig:grad_vis}). The gradient responses primarily come from the pixels not confidently predicted and correspond to the cues used for hemorrhage prediction. Fig.~\ref{fig:grad_vis} shows that FCN captures long range dependencies that can easily overfit to limited data, while PatchFCN focuses on the local morphology and may generalize better.

\vspace{-2mm}
\subsubsection{Patch-based Sliding Window Inference:} 
At inference time, standard FCN applies on the whole image at once \cite{long2015fully}. We hypothesize that this is sub-optimal for PatchFCN because the model is only trained on patches but has to take whole images at test time. That is why the default PatchFCN adopts sliding window inference to minimize the domain shift by letting PatchFCN evaluate patch by patch at test time. In Table \ref{table:sliding_window}, we show that sliding window inference consistently improves over fully convolutional inference for all patch sizes.  Note that the gap is largest for the smallest crop size of $80$, and decreases as patch size increases.

\begin{table}[t]
\vspace{-2mm}
     \centering
     \def\arraystretch{1.1}\tabcolsep=6pt
     \begin{tabular}{| c c c | c c c c |} 
     \hline
     $C$ & $B$ & Epoch & Dice & Jaccard & PixelAP & FrameAP \\
     \hline
     80 & 144 & 3600 & 69.4 (\textcolor{red}{-6.1}) & 53.1 (\textcolor{red}{-7.6})& 74.9 (\textcolor{red}{-3.6})& 85.5 (\textcolor{red}{-2.3})\\
     \hline
     120 & 64 & 1600 & 75.0 (\textcolor{red}{-0.9}) & 60.0 (\textcolor{red}{-1.2})& 75.6 (\textcolor{red}{-2.5})& 88.7 (\textcolor{red}{-0.6})\\
     \hline
     240 &  16 & 400 & 75.9 (\textcolor{red}{-0.7}) & 61.2 (\textcolor{red}{-0.8})& 76.4 (\textcolor{red}{-2.1})& 89.8 (\textcolor{red}{-0.1})\\
     \hline
    \end{tabular}
    \vspace{-1mm}
    \caption{Sliding window inference consistently outperforms fully convolutional inference (black numbers) for all patch sizes. The red numbers show the gap with sliding window inference.}    
    \label{table:sliding_window}
    \vspace{-8mm}
\end{table}

\vspace{-5mm}
\section{Conclusion}
\vspace{-2mm}
We propose PatchFCN -- a simple yet effective framework for intracranial hemorrhage detection. PatchFCN approaches the performance of an expert neuroradiologist as well as performs competitively with the state-of-the-art at stack level. In addition, it localizes many subtypes of hemorrhages well and has strong pixel level performance. Analyses show that PatchFCN outperforms FCN by finding a good trade-off between batch diversity and the amount of context. Our work shows the capability of PatchFCN for intracranial hemorrhage detection and potentially for other medical segmentation tasks.

{\small
\bibliographystyle{splncs}
\bibliography{egbib}

\begin{thebibliography}{10}

\bibitem{arbabshirani2018advanced}
Arbabshirani, M.R., Fornwalt, B.K., Mongelluzzo, G.J., Suever, J.D., Geise,
  B.D., Patel, A.A., Moore, G.J.:
\newblock Advanced machine learning in action: identification of intracranial
  hemorrhage on computed tomography scans of the head with clinical workflow
  integration.
\newblock npj Digital Medicine \textbf{1}(1) (2018) ~9

\bibitem{titano2018automated}
Titano, J.J., Badgeley, M., Schefflein, J., Pain, M., Su, A., Cai, M.,
  Swinburne, N., Zech, J., Kim, J., Bederson, J.,  et~al.:
\newblock Automated deep-neural-network surveillance of cranial images for
  acute neurologic events.
\newblock Nat Med \textbf{24}(9) (2018)  1337--1341

\bibitem{lee2019explainable}
Lee, H., Yune, S., Mansouri, M., Kim, M., Tajmir, S.H., Guerrier, C.E., Ebert,
  S.A., Pomerantz, S.R., Romero, J.M., Kamalian, S.,  et~al.:
\newblock An explainable deep-learning algorithm for the detection of acute
  intracranial haemorrhage from small datasets.
\newblock Nature Biomedical Engineering \textbf{3}(3) (2019)  173

\bibitem{chang2018hybrid}
Chang, P., Kuoy, E., Grinband, J., Weinberg, B., Thompson, M., Homo, R., Chen,
  J., Abcede, H., Shafie, M., Sugrue, L.,  et~al.:
\newblock Hybrid 3d/2d convolutional neural network for hemorrhage evaluation
  on head ct.
\newblock American Journal of Neuroradiology \textbf{39}(9) (2018)  1609--1616

\bibitem{long2015fully}
Long, J., Shelhamer, E., Darrell, T.:
\newblock Fully convolutional networks for semantic segmentation.
\newblock In: CVPR. (2015)

\bibitem{zhang2018deep}
Zhang, Y., Chung, A.C.:
\newblock Deep supervision with additional labels for retinal vessel
  segmentation task.
\newblock In: International Conference on Medical Image Computing and
  Computer-Assisted Intervention, Springer (2018)  83--91

\bibitem{qin2018autofocus}
Qin, Y., Kamnitsas, K., Ancha, S., Nanavati, J., Cottrell, G., Criminisi, A.,
  Nori, A.:
\newblock Autofocus layer for semantic segmentation.
\newblock In: International Conference on Medical Image Computing and
  Computer-Assisted Intervention, Springer (2018)  603--611

\bibitem{wang2017multi}
Wang, H., Moradi, M., Gur, Y., Prasanna, P., Syeda-Mahmood, T.:
\newblock A multi-atlas approach to region of interest detection for medical
  image classification.
\newblock In: International Conference on Medical Image Computing and
  Computer-Assisted Intervention, Springer (2017)  168--176

\bibitem{zhang2018convolutional}
Zhang, Y., Yu, H.:
\newblock Convolutional neural network based metal artifact reduction in x-ray
  computed tomography.
\newblock IEEE transactions on medical imaging \textbf{37}(6) (2018)
  1370--1381

\bibitem{karayumak2018harmonizing}
Karayumak, S.C., Kubicki, M., Rathi, Y.:
\newblock Harmonizing diffusion mri data across magnetic field strengths.
\newblock In: International Conference on Medical Image Computing and
  Computer-Assisted Intervention, Springer (2018)  116--124

\bibitem{yu2017dilated}
Yu, F., Koltun, V., Funkhouser, T.:
\newblock Dilated residual networks.
\newblock In: CVPR. (2017)

\end{thebibliography}
}

\end{document}